\documentclass[10pt,twocolumn,letterpaper]{article}

\usepackage{cvpr}              

\usepackage{graphicx}
\usepackage{amsmath}
\usepackage{amssymb}
\usepackage{booktabs}

\usepackage{amsfonts}
\usepackage{float}
\usepackage{array}
\usepackage{color}
\usepackage{tabularx}
\usepackage{longtable}
\usepackage{tabu}
\usepackage{multirow}
\usepackage{url}
\usepackage{bm}
\usepackage[noend]{algpseudocode}
\usepackage{algorithmicx,algorithm}
\usepackage{adjustbox}
\usepackage{makecell}
\usepackage{diagbox}

\DeclareUnicodeCharacter{2212}{\textendash}

\usepackage[accsupp]{axessibility}  

\definecolor{grey}{RGB}{130,130,130} 
\definecolor{black}{RGB}{0,0,0}

\usepackage[pagebackref,breaklinks,colorlinks]{hyperref}

\usepackage[capitalize]{cleveref}
\crefname{section}{Sec.}{Secs.}
\Crefname{section}{Section}{Sections}
\Crefname{table}{Table}{Tables}
\crefname{table}{Tab.}{Tabs.}

\begin{document}

\title{Dual-Branch Network for Portrait Image Quality Assessment}

\author{Wei Sun$^{1}$, Weixia Zhang$^{1}$\thanks{Corresponding authors.}, Yanwei Jiang$^{1}$, Haoning Wu$^{2}$, Zicheng Zhang$^{1}$, \\
Jun Jia$^{1}$, Yingjie Zhou$^{1}$, Zhongpeng Ji$^{3}$, Xiongkuo Min$^{1}$, Weisi Lin$^{2}$, Guangtao Zhai$^{1}$\footnotemark[1]\\
$^1$Shanghai Jiao Tong University, $^2$Nanyang Technological University, $^3$Huawei
}

\maketitle

\begin{abstract}
Portrait images typically consist of a salient person against diverse backgrounds. With the development of mobile devices and image processing techniques, users can conveniently capture portrait images anytime and anywhere. However, the quality of these portraits may suffer from the degradation caused by unfavorable environmental conditions, subpar photography techniques, and inferior capturing devices. In this paper, we introduce a dual-branch network for portrait image quality assessment (PIQA), which can effectively address how the salient person and the background of a portrait image influence its visual quality. Specifically, we utilize two backbone networks (\textit{i.e.,} Swin Transformer-B) to extract the quality-aware features from the entire portrait image and the facial image cropped from it. To enhance the quality-aware feature representation of the backbones, we pre-train them on the large-scale video quality assessment dataset LSVQ and the large-scale facial image quality assessment dataset GFIQA. Additionally, we leverage LIQE, an image scene classification and quality assessment model, to capture the quality-aware and scene-specific features as the auxiliary features. Finally, we concatenate these features and regress them into quality scores via a multi-perception layer (MLP). We employ the fidelity loss to train the model via a learning-to-rank manner to mitigate inconsistencies in quality scores in the portrait image quality assessment dataset PIQ. Experimental results demonstrate that the proposed model achieves superior performance in the PIQ dataset, validating its effectiveness. The code is available at \url{https://github.com/sunwei925/DN-PIQA.git}.
\end{abstract}

\section{Introduction}
\label{sec:intro}
Image quality assessment~\cite{zhai2020perceptual} (IQA) plays an important role in various image processing systems. Over the past two decades, numerous IQA studies have been conducted to investigate image quality from both the subjective and objective perspectives. Subjective IQA studies~\cite{mantiuk2012comparison} normally designed a human study to collect the ground truth (\textit{i.e.} mean opinion scores, etc.) of the image quality. Typical subjective IQA methods include single stimulus, double stimulus, stimulus comparison, etc.~\cite{antkowiak2000final} Objective IQA studies utilize the computational models to compute the image quality aligned with human perception, which are further divided into three categories: full-reference (FR)~\cite{zhang2012comprehensive}, reduced-reference (RR)~\cite{dost2022reduced}, and no-reference (NR)~\cite{kamble2015no} based on whether reference images are required. The NR IQA models are also known as blind image quality assessment (BIQA).

\begin{figure*}[h]
    \centering
    \includegraphics[width=0.8\linewidth]{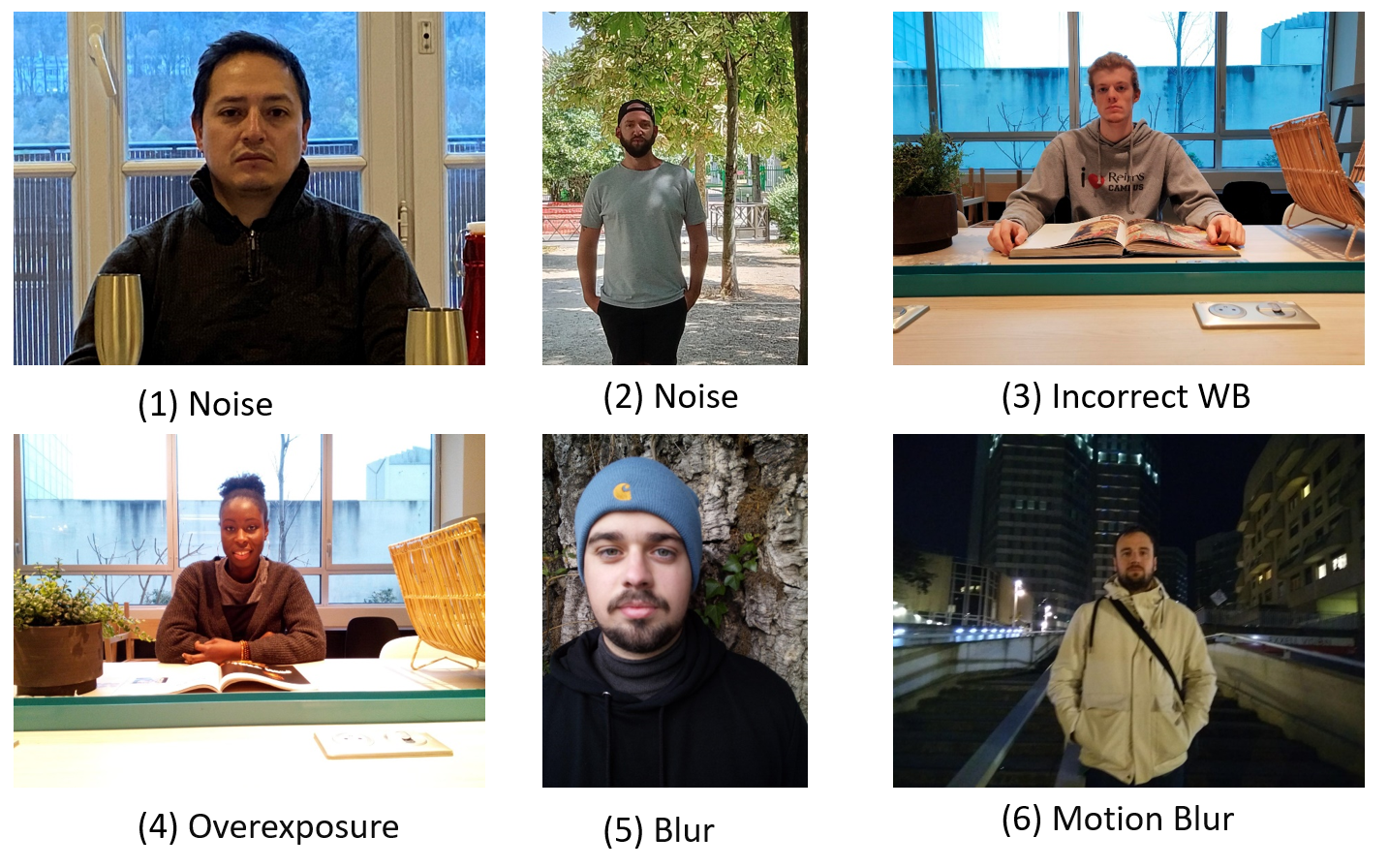}
    \caption{Some example of typical portrait image distortions.}
    \label{fig:1_dist}
\end{figure*}

In literature, most IQA models are designed for natural scene images (NSIs), including both synthetically distorted NSIs~\cite{moorthy2011blind} and authentically distorted NSIs~\cite{sun2023blind}. For example, natural scene statistics (NSS)-based IQA methods like BRISQUE~\cite{mittal2012no} and NIQE~\cite{mittal2012making} utilize the statistics characteristics of natural scene images to evaluate their quality. However, with the development of camera devices and image generation techniques, various types of images such as portrait images~\cite{chahine2023image}, HDR images~\cite{hanhart2015benchmarking}, VR images~\cite{sun2019mc360iqa,sun2018large,duan2023attentive}, gaming images~\cite{wang2022deep,wang2021multi}, point cloud~\cite{zhang2022no,zhang2022mm,zhang2023evaluating}, etc., have emerged. These images may possess characteristics significantly different from natural scene images, necessitating specific IQA methods for evaluating their quality.

In this paper, we investigate the portrait image quality assessment (PIQA) problem. Portrait image is a kind of photography style that primarily focuses on the face and expression of subjects, while minimizing emphasis on the background. Nowadays, the camera systems~\cite{delbracio2021mobile} of most mobile devices provide the portrait model to capture high-quality portrait images by simulating the shallow depth of field effect commonly achieved with professional cameras and prime. Specifically, the depth sensing module of mobile devices utilizes dual or multiple camera setups, in conjunction with specialized sensors like depth sensors or time-of-flight (ToF) sensors to capture depth information for distinguishing the human face and the background. Subsequently, the camera imaging system analyzes the scene to identify the person, ensuring it remains in focus while blurring the background. Additionally, the camera imaging system may employ image enhancement techniques such as skin smoothing, facial contouring, and lighting adjustments to enhance the overall appearance of the portrait.

During the portrait imaging process, various distortions may be introduced to the portrait image, thus degrading its perceptual quality. For example, the imaging system must accurately distinguish between the person and the background to effectively enhance the person while blurring the background. If the imaging system fails to do it, the person may also be blurred, leading to a poor image. The lighting also has a significant impact on portrait image quality. Extremely low-light environments prompt the camera to increase the gain of luminance channel in images, thereby amplifying the noise. Conversely, overly bright environments may cause background areas to be overexposed. Moreover, diverse image enhancement methods integrated into imaging systems may introduce issues like color shift, over-enhancement, etc., which still be a challenging problem for the IQA field. We list some typical portrait image distortions in Figure \ref{fig:1_dist}.

Recently, some works have tried to study the portrait image quality assessment problem. Chahine \textit{et al.}~\cite{chahine2023image} first investigate the PIQA problem subjectively. They construct a large-scale portrait IQA dataset PIQ, consisting of $5,116$ images of $50$ predefined scenarios by $100$ mobile phones, and then perform a human study to annotate the quality label of each image through the pairwise comparison method. Based on PIQ, they develop a semantic-aware BIQA portrait IQA method SEM-HyperIQA~\cite{chahine2023image}, which employs Hyper-IQA to capture the semantic information and utilizes the multi-task learning to derive the quality scores and image scene types, thus enforcing the model to perform scene-specific rescaling. They then introduce FHIQA~\cite{chahine2024generalized}, a PIQA method that also introduces a quality score rescaling method based on image semantics, to improve the ability of fine-grained image quality assessment and ensure robust generalization ability. They~\cite{chahine2024picniq} further utilize a siamese netowrk (\textit{i.e.,} ResNet-50) to extract quality-aware features from two different portrait images and employ a hub layer for pairwise comparisons training.

Although significant efforts have been made to explore the PIQA problem, there is still room to develop a more effective PIQA method. First, previous PIQA methods~\cite{chahine2023image,chahine2024picniq} typically involve cropping the high-resolution portrait image into several small-scale (\textit{i.e.,} $224\times224$) patches and then extracting the semantic and quality-aware features from these patches. It is important to note that the visual quality of portrait images is strongly influenced by both the person's face and their background. However, small patches cannot may not effectively capture complete semantic information and also may not adequately explore the relationship of how face and background features interactively influence the quality of portrait images. Second, the person and scene diversity in the current PIQA dataset is relatively limited, potentially hindering the training of a robust PIQA model capable of generalizing across various subjects and capturing scenes. Third, the current PIQA dataset PIQ~\cite{chahine2023image} utilizes the pairwise comparison method to assess the relative quality between two portrait images and calculate the just-objectionable-difference (JOD) as the ground truth of quality scores. Therefore, the JOD scores are only meaningful when comparing two portraits from the same scene. While previous studies have attempted the scene-specific rescaling and sparse comparison training methods to train the PIQA model, we demonstrate that a simpler learning-to-rank training method can achieve better performance.

To address the above challenges, we propose a dual-branch neural network for portrait image quality assessment. The proposed PIQA model consists of two branch networks: one is dedicated to extracting the semantic and quality-aware features from the entire portrait image, while the other is responsible for extracting the corresponding features from the facial image cropped from the portrait. This approach allows the proposed model to consider both the influence of the person and the background on the visual quality. To further encode the image into scene and quality-aware features, we employ LIQE~\cite{zhang2023blind}, a CLIP~\cite{radford2021learning} based image scene classification and image quality assessment model, to encode the portrait image into scene-aware and quality-aware features. 

Subsequently, we concatenate features from the entire image and facial features and LIQE feature and perform a two-layer multi-layer perception (MLP) to regress these features into the image score. To improve the model robustness and generalization, we pre-train the two branch networks on LSVQ~\cite{ying2021patch} and GFIQA~\cite{su2023going} datasets respectively. LSVQ is a large-scale video quality assessment (VQA) dataset consisting of $38,811$ video collected from the Internet. We choose the VQA dataset for pre-training because, compared to IQA datasets, it can provide more diverse scenes, complex distortions, and a larger number of frames for training, thus allowing to train a more powerful quality assessment model. GFIQA contains $20,000$ human faces along with the quality labels, thus enabling the model to learn a quality-aware representation for facial images. During the training, we randomly sample pairs of portrait images from the same scene and employ the fidelity loss to optimize the model. Experimental results demonstrate that the proposed model achieves the best performance on the portrait image quality assessment dataset PIQ~\cite{chahine2023image}, validating its effectiveness.

\section{Related Work}
\label{sec:related_work}
Since the portrait image quality assessment primarily focuses on face quality, this section reviews face-related IQA studies.

\subsection{Face-related IQA Datasets}
Previous face image datasets including LFW~\cite{huang2008labeled}, VGGFace2~\cite{cao2018vggface2}, CASIA-WebFace~\cite{yi2014learning}, etc., are primarily designed for face recognition and do not include the perceptual quality labels. 
While for general IQA datasets, such as LIVE~\cite{sheikh2006statistical}, CSIQ~\cite{larson2010most}, TID2013~\cite{ponomarenko2015image}, KADID-10k~\cite{lin2019kadid}, KonIQ-10k~\cite{hosu2020koniq}, SPAQ~\cite{fang2020perceptual} etc., they normally focus on the visual quality of natural scene images with synthetic or authentic distortions rather than solely focusing on human faces.
To investigate the problem of face-related image quality assessment, several subjective quality assessment studies have been conducted. For example, Liao \textit{et al.}~\cite{liao2012facial} construct a large-scale IQA dataset including $22,720$ various facial images, each of which was rated by $10$ individuals on a five-point quality scale. Su \textit{et al.}~\cite{su2023going} create GFIQA, a large-scale face IQA dataset comprising $20,000$ face images with authentic distortions, with each face corresponding to an individual person. Liu \textit{et al.}~\cite{liu2024assessing} establish the FIQA dataset, which contains of $42,125$ face images corrupted by diverse types of distortions, including single distortions (\textit{e.g.,} compression, blur, noise, etc.), enhancement algorithm-introduced distortions, and mixed distortions, etc. Chahine \textit{et al.} develop PIQ~\cite{chahine2023image}, a portrait IQA dataset consisting of $5,116$ single portrait images labeled by experts across three quality dimensions: overall, details and exposure. Importantly, the quality scores in PIQ are derived solely through comparative evaluations. 

Besides these 2D face IQA studies, increasing attention has been paid to quality assessment of 3D human faces\footnote{They are also called digital faces or digital humans.} and talking heads. For instance, Zhang \textit{et al.} have introduced several quality assessment datasets including DHHQA \cite{dhhqa}, SJTU-H3D \cite{h3d}, and DDH-QA \cite{ddhqa}, focusing on 3D digital human faces, static and dynamic 3D digital humans, respectively. Zhang \textit{et al.}~\cite{zhang2024comparative} investigate the quality of talking head videos and thus construct a dataset consisting of videos that were generated by audio-driven talking head generation algorithms. Zhou \textit{et al.}~\cite{zhou2024thqa} create the THQA dataset, rated $800$ talking videos that were generated by eight speech-driven methods by $40$ subjects.

\begin{figure*}[t]
    \centering
    \includegraphics[width=\textwidth]{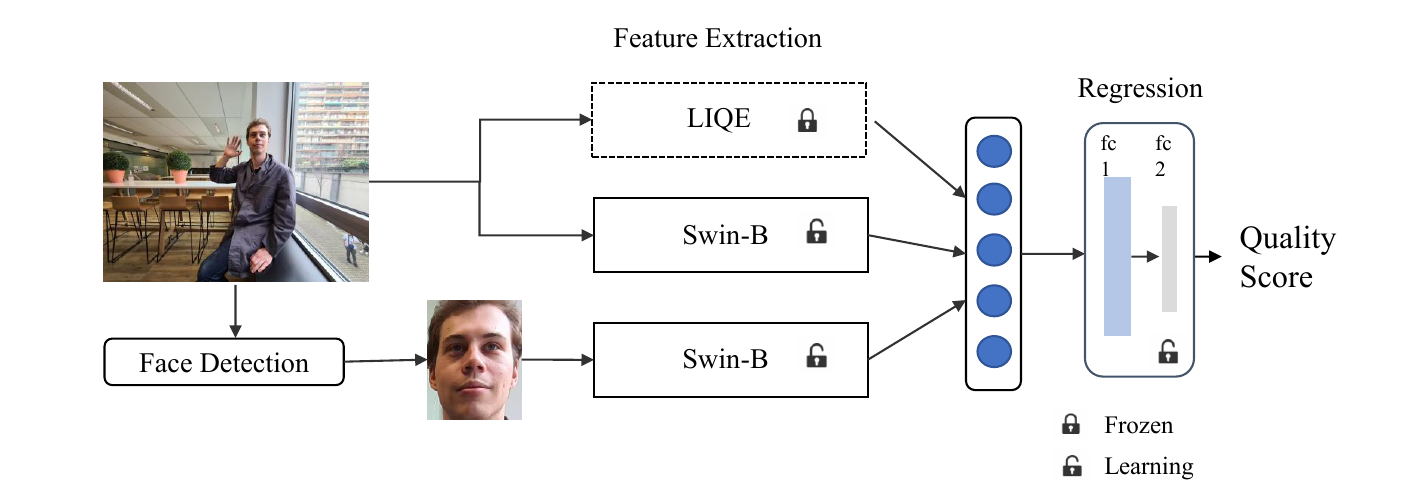}
    \caption{The framework of the proposed model. It consists of two branches, where one branch is used to extract the full image features and the other is utilized to extract the facial image features. We further extract the LIQE features as the auxiliary features to enhance the feature representation capabilities.}
    \label{fig:2_model}
\end{figure*}

\subsection{Face-related IQA Methods}
In past two decades, many general IQA methods have been proposed, including knowledge-driven methods~\cite{moorthy2011blind,saad2012blind,mittal2012no,zhai2019free,gu2014using,zhai2021perceptual,zhang2022no,sun2020dynamic} and data-driven methods~\cite{kang2014convolutional,bosse2017deep,ma2017end,zhang2018blind,su2020blindly,zhu2021generalizable,ke2021musiq, sun2019mc360iqa, lu2021cnn,sun2022deep,sun2023blind,lu2022deep,lu2022blind,wang2022deep,sun2022deep,sun2021deep,zhang2023md}. While these methods have demonstrated promising performance in assessing the quality of general images, they still lack sufficient capability to evaluate the quality of facial images. Therefore, some studies have been dedicated to developing effective IQA methods for facial images. We divide them into utility-oriented face IQA methods and fidelity-oriented face IQA methods.

\noindent\textbf{Utility-oriented face IQA methods} are used to evaluate whether a face image is suitable for face recognition systems. Some studies~\cite{sang2009face,wasnik2017assessing,khodabakhsh2019subjective} primarily focus on the effect of specific factors such as illumination, blur, noise, head pose, etc., on the quality of face recognition quality. 
For example, Wasnik \textit{et al.}\cite{wasnik2017assessing} develop a facial image quality metric based on vertical edge density, which can robustly estimate pose variations and enhance the quality assessment of a face image with different head poses. 
Khodabakhsh \textit{et al.}\cite{khodabakhsh2019subjective} analyze the evaluation effect of different kinds of objective face IQA methods as well as subjective scores provided by human subjects for face recognition systems using facial images captured under varying head poses, illumination conditions, and distances. 
Recently, some studies~\cite{grother2020ongoing,lijun2019multi,hernandez2019faceqnet,ou2021sdd,jo2023ifqa,boutros2023cr} propose more general face quality assessment metrics for in-the-wild facial images to assist face recognition. For example, Hernandez \textit{et al.} propose FaceQnet~\cite{hernandez2019faceqnet}, which selects the highest quality images as a gold-standard reference and computes the Euclidean distance between distorted images and the golden images in the features spaces as the quality score. 
Boutros \textit{et al.}~\cite{boutros2023cr} propose CR-FIQA, which evaluates the face quality by predicting its relative classifiability. 
This classifiability is then assessed based on the allocation of the feature representation of the training sample in angular space relative to its class center and the nearest negative class center. 
Kim \textit{et al.}~\cite{kim2024ig} develop IG-FIQA, a method aimed at addressing the issue where the quality scores used as pseudo-labels assigned from images of classes with low intra-class variance may not be unrelated to the actual quality. They thus propose a weight parameter to alleviate this negative impact.

\noindent\textbf{Fidelity-oriented face IQA methods} aim to evaluate the perceptual visual quality of facial images. Liao \textit{et al.}~\cite{liao2012facial} extract the Gabor features of facial images and then employ a hierarchical binary decision tree-based SVM to train a face quality evaluator. Su \textit{et al.}~\cite{su2023going} develop a face IQA model by leveraging generative priors. They extract intermediate generative features of images as latent references corresponding to the distorted target images to enhance the quality evaluation performance. Liu \textit{et al.}~\cite{liu2024assessing} introduce TransFQA, which uses the facial component guided Transformer network to integrate the global context, face region, and facial component detail features via the progressive attention mechanism. What's more, a distortion-specific prediction network is employed to derive the quality scores. For portrait IQA, Chahine \textit{et al.} introduce SEM-HyperIQA~\cite{chahine2023image}, which utilizes Hyper-IQA to capture the semantic information and employs the multi-task learning to derive the quality scores and image scene types. This approach enforces the model to perform scene-specific rescaling. Chahine \textit{et al.} develop FHIQA~\cite{chahine2024generalized}, which performs a quality score rescaling method based on image semantics to enhance fine-grained image quality assessment and ensure robust generalization ability. Chahine \textit{et al.}~\cite{chahine2024picniq} further use a Siamese network to extract quality-aware features of two different portrait images and utilize a hub layer for pairwise comparisons training. For digital human face, Zhou \textit{et al.} propose a no-reference quality assessment approach based on multi-task learning for the DHHQA dataset \cite{vitqa}, while Zhang \textit{et al.} pioneer quality assessment methods for 3D DH by integrating 3D features, action features, and prompts \cite{zhang2023geometry,chen2023no,zhang2024reduced}.

\section{Proposed Model}
\label{sec:method}
In this section, we introduce the proposed dual-branch PIQA model in detail. Our model is motivated by the observation that the quality of portrait images is primarily influenced by both the facial image region and the background region. Therefore, we perform a dual-branch network to extract the quality-aware features from the facial image and the entire image respectively. Additionally, we leverage a pre-trained model (\textit{i.e.,} LIQE~\cite{zhang2023blind}) for image quality assessment and scene classification model to extract the auxiliary quality-ware and scene-specific features to enhance the proposed model.

\subsection{Dual-branch Network}
As depicted in Figure \ref{fig:2_model}, we design a two-branch network structure for portrait image quality assessment. Our framework is very simple, comprising only two Swin Transformer-B~\cite{liu2021swin} as the feature extractors, where one branch processes the entire image and the other focuses on the facial image. We do not adopt the commonly used Siamese structure in previous IQA studies because we think the facial images and the entire images exhibit distinct characteristics and require different model weights to extract the optimal quality-aware features. Moreover, as an effective optimization approach, we pre-train the two feature extractors on the public I/VQA datasets to enhance the feature extraction capabilities. Specifically, for Swin Transformer-B processing the entire image as input, we train it on LSVQ~\cite{ying2021patch} using the method in~\cite{sun2023analysis} to obtain the general quality-aware feature extractor. For the other Swin-Transformer-B, we train it on GFIQA~\cite{su2023going} to obtain the facial quality-aware feature extractor. We formulate this procedure as follows:

\begin{equation}
\begin{aligned}
\mathcal{F}_{full} &= {\rm Swin}_{full} (\bm x_{full}), \\
\mathcal{F}_{facial} &= {\rm Swin}_{facial} (\bm x_{facial}), \\
\end{aligned}
\end{equation}
where $x_{full}$ and $x_{facial}$ are the entire portrait image and the facial image cropped from the entire image using the face detection method YOLO5Face~\cite{qi2022yolo5face}, respectively. $\mathcal{F}_{full}$ and $\mathcal{F}_{facial}$ are the quality-aware features of the entire portrait image and the facial image extracted by the feature extractors ${\rm Swin}_{full}$ and ${\rm Swin}_{facial}$ respectively.

\subsection{LIQE Features}
LIQE~\cite{zhang2023blind} is a visual-language model for image quality assessment and scene classification. It utilizes CLIP, consisting of an image encoder and a text encoder, to compute the cosine similarity between text features and image features. For LIQE, it takes a text prompt $\bm t(s, d, c) = $ ``\textit{a photo of a(n) \{s\} with \{d\} artifacts, which is of \{c\} quality}" along with an image as the inputs and calculate the cosine similarity between text features and image features to represent the probability of how well the text prompt describes the tested image. So, these probabilities can be utilized to infer the scene type, artifact type, and quality level of the tested image.

In this section, we use the probabilities derived from various types of text prompts as auxiliary features to characterize the scene, artifact, and quality-level attributes of video frames. We consider nine scene categories: $s \in S =$ \{“animal”, “cityscape”, “human”, “indoor scene”, “landscape”, “night scene”, “plant”, “still-life”, and “others”\}, eleven distortion types: $d \in D =$ \{“blur”, “color-related”, “contrast”, “JPEG compression”, “JPEG2000 compression”, “noise”, “overexposure”, “quantization”, “under-exposure”, “spatially-localized”, and “others”\}, and five quality levels: $c \in C = \{1, 2, 3, 4, 5\} =$ \{“bad”, “poor”, “fair”, “good”, “perfect”\}. Thus, we have a total of $495$ text prompt candidates to compute the probabilities:

\begin{equation}
\begin{aligned}
\mathcal{F}_{\rm LIQE} &= {\rm LIQE} (\bm {x_{full}, t(s,d,c)})), \\
\end{aligned}
\end{equation}
where $\mathcal{F}^{\rm LIQE}_i$ represents the LIQE features, which contains $495$ dimensions corresponding to the scene category, artifact type, and quality level characteristics.

\begin{table*}[h]
\caption{The performance of the proposed model and competing BIQA models on the PIQ dataset }
\label{tab:experiments}
\begin{tabularx}{\textwidth}{c|XXXX|XXXX|XXXX} 
\hline
\multirow{2}{*}{\diagbox{Model}{Attribute}} & \multicolumn{4}{c|}{Overall} & \multicolumn{4}{c|}{Exposure} & \multicolumn{4}{c}{Details} \\
\cline{2-13}
 & SRCC & PLCC & KRCC & MAE & SRCC & PLCC & KRCC & MAE & SRCC & PLCC & KRCC & MAE \\
\hline
DB-CNN (CIVE C) & 0.59 & 0.64 & 0.43 & 1.04 & 0.69 & 0.69 & 0.51 & 0.91 & 0.59 & 0.51 & 0.45 & 0.99 \\
MUSIQ (KonI0-10k) & 0.76 & 0.75 & 0.57 & 0.95 & 0.74 & 0.70 & 0.55 & 0.93 & 0.71 & 0.67 & 0.52 & 0.88 \\
MUSIQ (PaQ-2-PiQ) & 0.74 & 0.74 & 0.54 & 1.09 & 0.79 & 0.78 & 0.59 & 0.87 & 0.72 & 0.77 & 0.53 & 0.90 \\
HyperIOA & 0.74 & 0.74 & 0.55 & 0.99 & 0.69 & 0.68 & 0.50 & 0.86 & 0.70 & 0.67 & 0.50 & 0.94 \\
SEM-HyperIQA & 0.75 & 0.75 & 0.56 & 1.03 & 0.72 & 0.70 & 0.53 & 0.97 & 0.73 & 0.65 & 0.55 & 0.88 \\
SEM-HyperIQA-CO & 0.74 & 0.74 & 0.55 & 1.04 & 0.70 & 0.70 & 0.52 & 0.94 & 0.75 & 0.71 & 0.55 & 0.85 \\
FHIQA & 0.78 & 0.78 & 0.59 & 1.12 & 0.76& 0.71 & 0.57 & 0.85 & 0.74 & 0.72 & 0.55 & 0.80\\
PICNIQ & 0.81 & 0.81 & 0.61 &  0.72& 0.77& 0.80&0.60 & 0.76 & 0.83 & 0.81&0.64  &0.72\\
StairIQA &0.72 & 0.75& 0.55& 0.70& 0.69& 0.73&0.51 &0.72 &0.67 &0.73 &0.50 &0.73 \\
LIQE &0.81 &0.80 &0.62 &0.64 &\textbf{0.85} &0.82 &0.66 &0.66 &0.85 &0.82 & 0.67&0.60 \\
Proposed &\textbf{0.85} & \textbf{0.86}& \textbf{0.68}& \textbf{0.53}&\textbf{0.85} &\textbf{0.86} &\textbf{0.68} &\textbf{0.53} & \textbf{0.86}& \textbf{0.87}& \textbf{0.69}&\textbf{0.52}  \\
\hline
\end{tabularx}
\centering
\end{table*}

\subsection{Quality Regression}
After calculating these features, we further concatenate these features into the final feature representation $\mathcal{F}$: 
\begin{equation}
\begin{aligned}
\mathcal{F} &= {\rm Cat}(\mathcal{F}_{full}, \mathcal{F}_{facial}, \mathcal{F}_{LIQE}), \\
\end{aligned}
\end{equation}
where $\rm Cat$ is the concatenation operator.

We then use a two-layer MLP network to regress $\mathcal{F}$ into the quality score:
\begin{equation}
\begin{aligned}
 \hat{q} &= {\rm MLP}(\mathcal{F}), \\
\end{aligned}
\end{equation}
where $\rm MLP$ denotes the MLP operator and $\hat{q}$ is the quality score of the portrait image.

\subsection{Loss Function}
Previous IQA studies normally utilize either L1 or L2 loss to minimize the discrepancy between model quality scores close to ground-truth quality scores during the training. However, the ground-truth quality scores (\textit{et al.} JODs) in PIQ are obtained through the pairwise comparison experiment, where only images from the same scenes were compared. It may cause inconsistent quality scales of images from different scenes. Therefore, we adopt the learning-to-rank training method, which can make the model learn the rank information of images from the same scene. 

In the training stage, we randomly sample two images ($\bm{x}$, $\bm{y}$) from the same scene and calculate the quality score of two images $\hat{q}(\bm{x})$ and $\hat{q}(\bm{y})$ through the proposed model. We then compute a binary label according to their ground-truth JODs:
\begin{equation}
    p(\bm{x}, \bm{y}) = \left\{ 
    \begin{aligned} 
    & 0 \quad {\rm{if}} \quad q(\bm{x}) \geq q(\bm{y})\\
    & 1 \quad \rm{otherwise}
    \end{aligned}
    \right.
    .
\end{equation}
where $q(\bm{x})$ and $q(\bm{y})$ are the ground-truth JODs of the image pair ($\bm{x}$, $\bm{y}$).
We estimate the probability of $\bm{x}$ perceived better than $\bm{y}$ as

\begin{equation}
\hat{p}(\bm{x}, \bm{y}) = \Phi(\frac{\hat{q}(\bm{x})-\hat{q}(\bm{y})}{\sqrt{2}}),
\end{equation}
where $\Phi(\cdot)$ is the standard Normal cumulative distribution function, and the variance is fixed to one. We adopt the fidelity loss~\cite{tsai2007frank} to optimize the model:
\begin{equation}
\begin{aligned}
\ell(\bm{x}, \bm{y};\bm{\theta}) = &1-\sqrt{p(\bm{x}, \bm{y})\hat{p}(\bm{x}, \bm{y})}\\
&-\sqrt{(1-p(\bm{x}, \bm{y}))(1-\hat{p}(\bm{x}, \bm{y}))}.
\end{aligned}
\end{equation}

\section{Experiment}
\subsection{Experimental Protocol}
\noindent\textbf{Test Dataset.} We test our model on the PIQ dataset~\cite{chahine2023image}, containing $5,116$ portrait images, where $3,630$ images are used for training and $1,486$ images are used for validating. A private test set is further used to test the performance of the proposed model for the NTIRE 2024 Portrait Quality Assessment ft. DXOMARK~\cite{chahine2024ntire_pqa}.\footnote{\url{https://codalab.lisn.upsaclay.fr/competitions/17311}}.

\vspace{0.2cm}
\noindent\textbf{Implementation Details.} As stated in Section \ref{sec:method}, we utilize Swin Transformer-B~\cite{liu2021swin} as the backbones. To improve the generalization ability of the backbone networks, we train the branch for the entire image on the LSVQ dataset~\cite{ying2021patch}, following the training strategy in~\cite{sun2023analysis}. Thus it can learn a general quality-aware feature representation ability. As for the branch for the facial image, we train it on the GFIQA dataset~\cite{su2023going}, enabling it can learn facial quality-aware feature representation. During the training stage, we resize the resolution of the minimum dimension of portrait images to $448$ while preserving their aspect ratios. During the training and test stages, the portrait images are randomly and centrally cropped with a resolution of 384$\times$384. The Adam optimizer with the initial learning rate $1\times10^{-5}$ and batch size $12$ is used to train the proposed model on a server with $2$ NVIDIA RTX 3090 GPUs. We decay the learning rate by a factor of $10$ after $2$ epochs and the total number of epochs is set as $10$.

\vspace{0.2cm}
\noindent\textbf{Compared Models.} We compare the proposed model with nine popular BIQA methods, including DB-CNN~\cite{zhang2018blind}, MUSIQ~\cite{ke2021musiq}, HyperIQA~\cite{su2020blindly}, SEM-HyperIQA~\cite{chahine2023image}, and SEM-HyperIQA-CO~\cite{chahine2023image}, FHIQA~\cite{chahine2024generalized}, PICNIQ~\cite{chahine2024picniq}, StairIQA~\cite{sun2023blind}, and LIQE~\cite{zhang2023blind}, where EM-HyperIQA, SEM-HyperIQA-CO, FHIQA, and PICNIQ are specifically designed for portrait images.

\vspace{0.2cm}
\noindent\textbf{Evaluation Criteria.}
We employ four criteria to evaluate the performance of VQA models: PLCC, Spearman rank-order correlation coefficient (SRCC), Kendall rank correlation coefficient (KRCC), and mean absolute error (MAE). Note that PLCC assesses the prediction linearity of the VQA model, SRCC and KRCC evaluate the prediction monotonicity, and MAE reflects prediction accuracy. An outstanding VQA model should achieve SRCC and PLCC values close to 1 and MAE close to 0.
Before computing PLCC, we adhere to the procedure outlined in~\cite{antkowiak2000final} to map model predictions to  MOSs by a monotonic four-parameter logistic function to compensate for prediction nonlinearity. We first calculate the values of four metrics for each individual scene and utilize the average results as the evaluation criteria.

\begin{table}
\small
\centering
\renewcommand{\arraystretch}{1.2}
\caption{The results on the private test set of NTIRE 2024 Portrait Quality Assessment Challenge}
\label{tab:challenge}
\begin{tabular}{c| ccc }
\toprule[.15em]
Team & SRCC & PLCC&  KRCC  \\
\hline
Xidian-IPPL& 0.554&0.597  &  0.381  \\
BDVQAGroup& 0.393&0.575  &  0.333  \\
SJTU MMLab (ours) & 0.411&  0.544&  0.333  \\
SECE-SYSU& 0.304&  0.453&0.238    \\
I²Group&  0.357&  0.433&  0.286  \\
\bottomrule[.15em]
\end{tabular}
\end{table}

\subsection{Experimental Results}
\begin{table}
\small
\centering
\renewcommand{\arraystretch}{1.2}
\caption{The results of ablation studies on the PIQ dataset}
\label{tab:abliation_experiment}
\begin{tabular}{ccc| cc }
\toprule[.15em]
\multirow{2}{*}{\makecell[c]{Full Image \\Branch}} & \multirow{2}{*}{\makecell[c]{Facial Image \\Branch}} & \multirow{2}{*}{LIQE}   & \multicolumn{2}{c}{Overall}  \\
 &  & &  SRCC & PLCC \\
\hline
$\surd$& $\times$& $\times$ & 0.82& 0.84  \\
$\times$& $\surd$& $\times$ &0.82 &  0.83 \\
$\surd$& $\surd$& $\times$ &0.84 &  0.85 \\
$\surd$& $\surd$& $\surd$ &0.85 & 0.86   \\
\bottomrule[.15em]
\end{tabular}
\end{table}

We present the experimental results in Table~\ref{tab:experiments}, from which we draw several conclusions. First, we observe that PICNIQ, LIQE, and the proposed models exhibit significant superiority over other BIQA models. This is attributed to their utilization of the learning-to-rank strategy for optimizing the BIQA models, whereas other methods employ either L2 or Huber Loss directly. The former approach effectively tackles the scaling inconsistency problem in PIQA, leading to better performance. Second, the proposed model achieves the highest performance among the compared BIQA methods and is also superior to the models SEM-HyperIQA, SEM-HyperIQA-CO, and PICNIQ, which are specifically developed for portrait images, indicating its effectiveness in evaluating the visual quality of portrait images. This demonstrates the rationality of our model structure in respectively extracting features from entire and facial images.

We list the results of NTIRE 2024 Portrait Quality Assessment Challenge in Table~\ref{tab:challenge}. From Table~\ref{tab:challenge}, it is shown that our model achieves second place in terms of SRCC and KRCC metrics, and third place in terms of PLCC metric, which further demonstrates the effectiveness of the proposed model.

\subsection{Ablation Studies}
In this section, we validate the effectiveness of the components of our model. Specifically, we train three variants of the proposed model: one consisting solely of entire image feature extraction, another consisting solely of facial image feature extraction, and a third consisting of both the full image and facial image feature extraction but without LIQE features. The experimental results are listed in Table~\ref{tab:abliation_experiment}. From Table~\ref{tab:abliation_experiment}, we observe that solely utilizing either full image extraction branch or facial image extraction results in subpar performance, which indicates that PIQA needs to consider both person and background characteristics. Removing the LIQE features causes a performance drop of $0.01$ in terms of both SRCC and PLCC, suggesting that auxiliary quality-aware and scene-specific features can promote the performance of PIQA.

\section{Conclusion}
In this paper, we develop a dual-branch network for portrait image quality assessment. We consider the PIQA problem as how to quantify the impact of both the facial image region and the background region on visual quality. To achieve this, we employ a dual-branch network to extract quality-aware features, where one branch is dedicated to extracting features from the entire image and the other branch focuses on extracting features from the facial image. To improve the capability of representing quality-aware features, we pre-train the two branches on large-scale VQA and facial IQA datasets, enabling them to adapt to diverse subjects and background scenes. Additionally, we incorporate LIQE features as auxiliary quality-aware and scene-specific features to further enhance the proposed model. Finally, we employ the fidelity loss function to address the label inconsistency problem in the PIQA dataset. Experimental results show that our model outperforms competing BIQA methods on the PIQ dataset by a large margin, demonstrating its effectiveness.

\section{Acknowledgement}
This work was supported in part by the National Natural Science Foundation of China under Grants 62071407, 62301316, 62371283, 62225112, 62376282 and 62271312, the China Postdoctoral Science Foundation under Grants 2023TQ0212 and 2023M742298, the Postdoctoral Fellowship Program of CPSF under Grant GZC20231618, the Fundamental Research Funds for the Central Universities, the National Key R\&D Program of China (2021YFE0206700), the Science and Technology Commission of Shanghai Municipality (2021SHZDZX0102), and the Shanghai Committee of Science and Technology (22DZ2229005).

{\small
\bibliographystyle{ieee_fullname}
\bibliography{PaperForReview}
}

\end{document}